\documentclass[runningheads]{llncs}
\usepackage{graphicx}
\usepackage{subfigure} 
\usepackage{amsmath}
\usepackage{amsfonts}
\usepackage{multirow}
\usepackage{tabu}
\usepackage{bbding}
\usepackage{diagbox}
\usepackage[colorlinks, linkcolor=blue, citecolor=blue]{hyperref}
\usepackage{url}
\usepackage{cite}  
\usepackage{arydshln}
\usepackage{enumitem}
\usepackage{subfloat}
\usepackage[misc]{ifsym}
\usepackage{mathrsfs}
\usepackage{color}
\usepackage{footnote}
\makesavenoteenv{table}
\usepackage[colorlinks, linkcolor=blue, citecolor=blue]{hyperref}
\begin{document}
\title{Deps-SAN: Neural Machine Translation with Dependency-Scaled Self-Attention Network}
%
%
\author{
Ru Peng\inst{1} 
\and Nankai Lin\inst{2}
\and Yi Fang\inst{2}
\and Shengyi Jiang\inst{3}
\and Tianyong Hao\inst{4}
\and Boyu Chen\inst{5}
\and Junbo Zhao\inst{1}\Letter
}
\authorrunning{Peng et al.}
%
\institute{
College of Computer Science and Technology, Zhejiang University, China
\and School of Information, Guangdong University of Technology, China 
\and School of Information Science and Technology, Guangdong University of Foreign Studies, China
\and School of Computer Science, South China Normal University, China
\and Institute of Health Informatics, University College London, United Kingdom
\\
\email{\{pengru709909347@gmail.com, neakail@outlook.com, fangyi@gdut.edu.cn, jiangshengyi@163.com, haoty@m.scnu.edu.cn, boyu.chen.19@ucl.ac.uk, j.zhao@zju.edu.cn}\}}
\toctitle{Lecture Notes in Computer Science}
\tocauthor{Authors' Instructions}
\maketitle              
\begin{abstract}
Syntax knowledge contributes its powerful strength in Neural machine translation (NMT) tasks.
Early NMT works supposed that syntax details can be automatically learned from numerous texts via attention networks.
However, succeeding researches pointed out that limited by the uncontrolled nature of attention computation, the NMT model requires an external syntax to capture the deep syntactic awareness.
Although existing syntax-aware NMT methods have born great fruits in combining syntax, the additional workloads they introduced render the model heavy and slow. 
Particularly, these efforts scarcely involve the Transformer-based NMT and modify its core self-attention network (SAN).
To this end, we propose a parameter-free, \textbf{Dep}endency-\textbf{s}caled \textbf{S}elf-\textbf{A}ttention \textbf{N}etwork (Deps-SAN) for syntax-aware Transformer-based NMT. 
A quantified matrix of dependency closeness between tokens is constructed to impose explicit syntactic constraints into the SAN for learning syntactic details and dispelling the dispersion of attention distributions. 
Two knowledge sparsing techniques are further integrated to avoid the model overfitting the dependency noises introduced by the external parser.
Experiments and analyses on IWSLT14 German-to-English and WMT16 German-to-English  benchmark NMT tasks verify the effectiveness of our approach.
\end{abstract}
\section{Introduction}
Syntax knowledge occupies a pivotal position in learning the context of translation, i.e., the deep semantic modelling of the sentence. 
Consequently, incorporating syntax knowledge has attracted massive attention from the neural machine translation (NMT) community.
Early works such as \cite{Choi2017ContextdependentWR,Zhang8031316, zhou-etal-2017-chunk, chen2017translation} assumed that the NMT model can draw out syntax knowledge from large-scale bilingual texts via attention networks \cite{Bahdanau-Neural,luong-etal-2015-effective, vaswani2017attention}.
However, were the facts really like that?
Shi et al. \cite{shi-etal-2016-string} and Li et al. \cite{Li2017Modeling} found that the NMT network failed to capture sufficient internal syntax details of a sentence. 
The rationale behind the above failure is that the NMT model requires a parse-task-specific training paradigm to recover the hidden syntax in sentences.
Going further, Yang et al. \cite{yang2019context} revealed that this limitation of syntactic distillation arose from a deficiency in the attention mechanism. 
That is, the self-attention network (SAN) is solely controlled by two trainable parameter matrices when modeling the correspondences of query and key vectors:
\begin{figure*}[!t]
\centering
\includegraphics[width=0.9\textwidth]{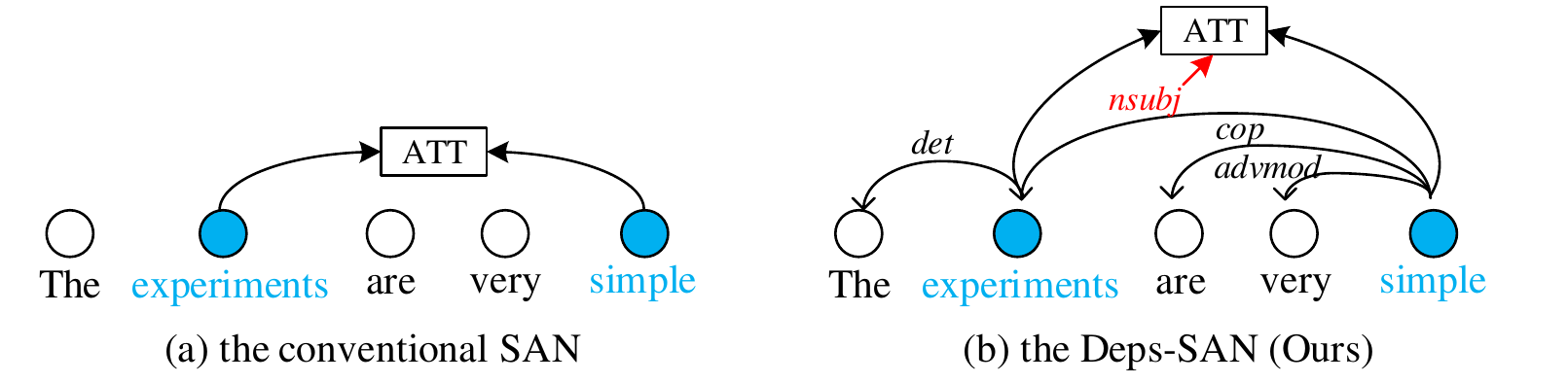}
\caption{Illustration of different attention network learning the correspondence between ``experiments'' and ``simple''.} \label{fig1}
\end{figure*}

\begin{equation}\label{eq:1}
Q{{K}^{T}}=(H{{W}_{Q}}){{(H{{W}_{K}})}^{T}}=H({{W}_{Q}}W_{K}^{T}){{H}^{T}},
\end{equation}
where $\left\{ {Q,K} \right\}$ are the query and key vectors, and $H$ is the hidden representation of input token. 
$\left\{ {{W}_{Q},{W}_{K}} \right\}$ are the trainable parameter matrices. 
From this viewpoint of mathematical theory, in the absence of external syntactic constraints, the attention network is trained for syntactic parsing in a weakly supervised manner.
To give an intuitive example in Fig. \ref{fig1}(a), the SAN individually computes the correspondence between ``experiments'' and ``simple'' without considering their syntactic dependency. 
Naturally, we expect that exploiting dependency syntax can guide the attention network to further improve the NMT performance, as done by Fig. \ref{fig1}(b).

Such being the case, let us revisit the existing research on syntax knowledge in NMT.
Recent promising syntax-aware MMT methods mostly focused on RNN-based architecture, which roughly fall into four categories: 
a) extra tree-RNN/CNN modules \cite{eriguchi-etal-2016-tree,chen2017improved,wu2017improved}; 
b) linearizing dependency tree \cite{sennrich-haddow-2016-linguistic,ma-etal-2018-forest,Ma2017Deterministic};
c) intermediate syntactic representations \cite{chen-etal-2017-neural,zhang-etal-2019-syntax};
d) multi-task with dependence parsing \cite{eriguchi2017learning,ShiWu-Dependency};
e) neural syntactic distance \cite{Peng2019NeuralMT,omote-etal-2019-dependency,ma2019improving};
e) syntax-aware data augmentation \cite{duan2020syntax}.
Although the great efforts these works have made, the additional syntactic workloads (i.e. modules, linearization, representations, etc) render the NMT model heavy and slow.
Moreover, the referred approaches scarcely explored to the more advanced Transformer-based NMT architecture\footnote[1]{Transformer, as the new generation of NMT baseline, abandons the recurrence and convolutions, and solely relies on SANs to achieve the incredible progress.} \cite{vaswani2017attention}, also do not modify the core SAN component by syntax information.

In response, we innovatively propose a parameter-free, \textbf{Dep}endency-\textbf{s}caled \textbf{S}elf-\textbf{A}ttention \textbf{N}etwork (Deps-SAN) for syntax-aware Transformer-based NMT.
Motivated by the uncontrolled attention weight in Eq. \ref{eq:1}, an idea that comes into our mind, we can inject the closeness of dependencies between tokens into the training of SANs to learn syntactic details. 
We enable this by proposing a dependency-scaled matrix employing the dependency distance of words derived from the parsed dependency tree.
From our experiments, we find that this quantified matrix has notably imposed strong syntactic constraints as well as dispelled the dispersion of attention distribution.
Besides, to tackle the problem of overfitting the dependency noises for our approach, we further partially modify the tuned attention weight by incorporating  
random sampling sparsing (RS-Sparsing) or k-value window sparsing (Wink-Sparsing).

Overall, our contributions are summarized as follows:
\begin{itemize}
\item[I.] To guide the attention distribution of unconstrained SANs, we contribute a Deps-SAN for Syntax-aware NMT, which introduces quantitative syntax dependencies for the computation of traditional SANs.
\item[II.] Two knowledge sparsing methods are proposed to prevent Deps-SAN from overfitting dependency noises, which sheds some light on the anti-overfitting for SAN-based methods.
\item[III.] Extensive experiments and analyses on two benchmark NMT datasets demonstrate the effectiveness of our system. 
\end{itemize}

\section{Approach}
In this section, we describe the methodological details of our approach. 
The overall framework of our Deps-SAN is depicted in Fig. \ref{fig3}. 
Specifically, applying our method to Transformer-based NMT is supposed to go through three steps: 1) deriving dependency-scaled matrix; 2) constructing Deps-SAN; and 3) adding knowledge sparsing.

\subsection{Dependency-Scaled Matrix} 
\begin{figure*}[!t]
\centering
\includegraphics[width=1.\textwidth]{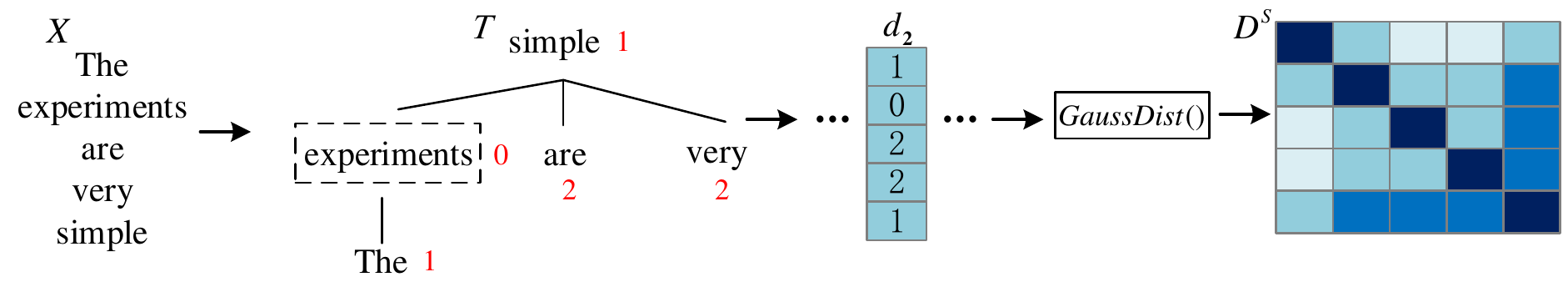}
\caption{The derivation pipeline of the dependency-scaled matrix.} \label{fig2}
\end{figure*}

To construct Deps-SAN, we first derive a dependency-scaled matrix to measure the closeness of syntactic dependencies, as shown in Fig. \ref{fig2}.
Given the input sentence $X$ with length $I$, in line with previous works \cite{wang2007binarizing,chen2018syntax,peng2021Syntax}, we extract the dependency tree $\mathcal{T}$ by an external syntax parser.
Next, we compute the word-level dependency distance based on the parsed dependency tree.
Specifically, we defined the dependency distance as the length of the path traversed from a word to another word on the tree, and the distance between two directly-connected words is assigned as 1.
For example, as displayed in Fig. \ref{fig2}, the dependency distance of ``experiments'' itself is 0, the dependency distance of ``simple'' and ``The'' is 1, and the dependency distance of ``are'' and ``very'' is 2.
In this way, the nearer the dependency distance between a word with the word ``experiments'', the closer the syntactic dependence between the word and the word ``experiments''. 
Then we traverse each word via the original word order and simultaneously count the dependency distance between the current traversed word and other words (including itself).
After that, we combine all dependency distance sequences ${d}_{i}$ and derive a dependency-scaled matrix ${{D}^{s}}\in {{\mathbb{R}}^{{IxI}}}$ by a Gaussian distribution. 
Each row of the matrix denotes the closeness of syntactic dependence of each word to other words.
As thus, when encoding the word ``experiments'', the SAN can discriminatively focus on words that are more correlated with their dependence with the help of the dependency-scaled matrix.

\subsection{Dependency-Scaled Self-Attention Network}
\begin{figure*}[!t]
\centering
\includegraphics[width=0.8\textwidth]{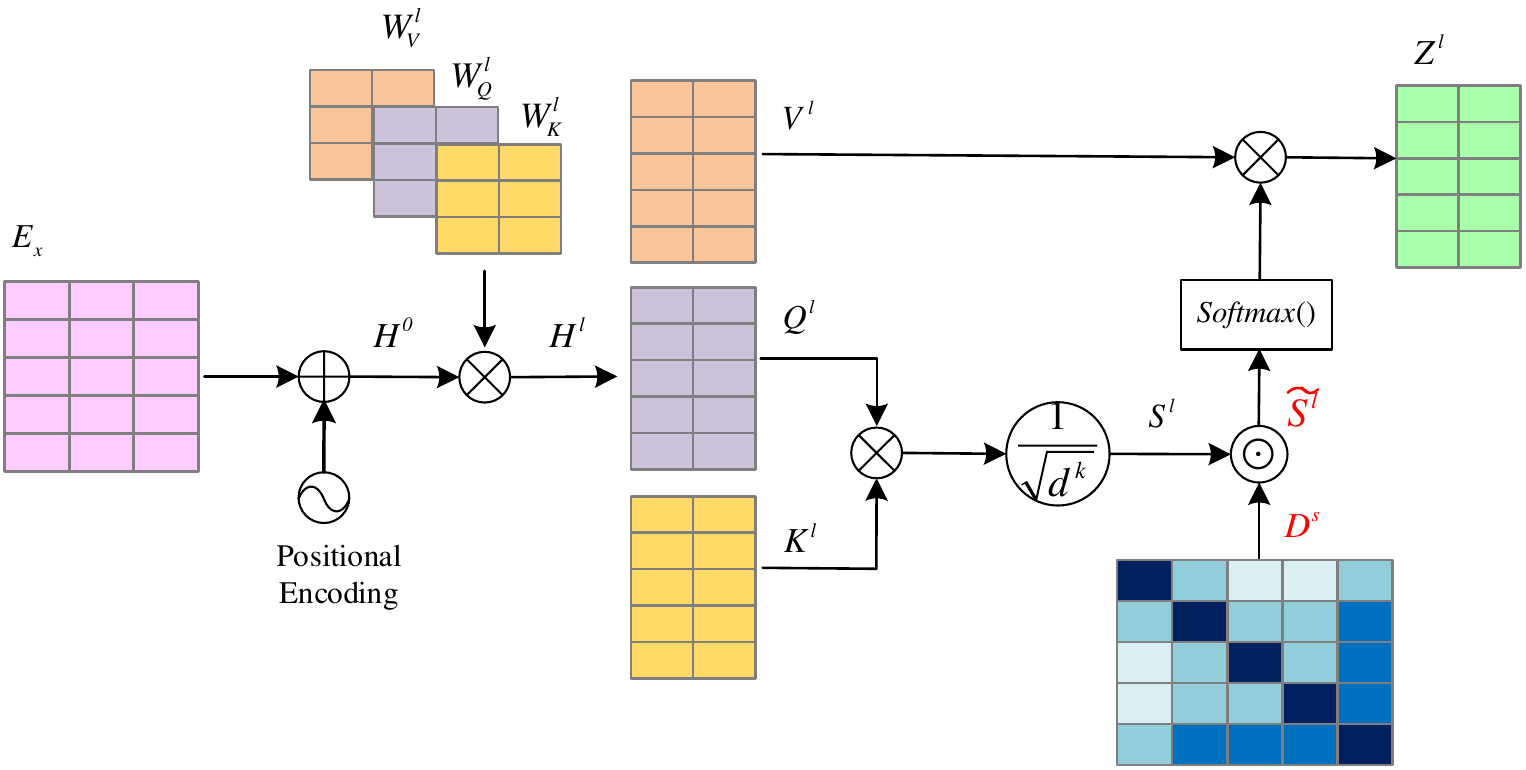}
\caption{The overall architecture of our Deps-SAN.} \label{fig3}
\end{figure*}

In this part, we will comprehensively introduce the overall architecture of Deps-SAN (i.e.  Fig. \ref{fig3}) and how to apply it to Transformer-based NMT.
For the source sentence $X$, the source annotation sequence $H$ was initialized by the sum of the word embeddings ${E}_{x}$ and the position encoding \cite{gehring2016convolutional}. 
Further, the dependency-scaled matrix ${D}^{s}$ and the source annotation sequence $H$ are both fed as the input of the $N$ attention heads in the Deps-SAN.
Analogous to the Transformer\footnote[2]{To distinctly illustrate our model, please refer to the original paper for details of Transformer-based NMT.} \cite{vaswani2017attention}, the query, key and value vectors of $l$-th encoder layer is calculated as follow:
\begin{equation}\label{eq:2}
\left\{ {{Q}^{l}},{{K}^{l}},{{V}^{l}} \right\}=\left\{ {H}^{l}W_{Q}^{l},{H}^{l}W_{K}^{l},{H}^{l}W_{V}^{l} \right\},
\end{equation}
where ${{Q}^{l}},{{K}^{l}},{{V}^{l}}\in {{\mathbb{R}}^{I\times {{d}_{k}}}}$ are three vectors that are linearly projected from the sentence annotation ${{H}^{l}}$ of the $l$-th encoder layer. 
Here, ${{d}_{k}}={{d}_{model}}/H$ and ${{H}^{0}}$ is the initial source annotation sequence.
Next, we compute the dot product between each query and all keys, and divide by $\sqrt{{d}_{k}}$ to obtain the alignment score ${S}^{l}$, which indicates how much attention should be placed on other words when annotating the current word. 
Following upon the alignment score, we explicitly impose syntactic constrains on this score by point-wise weighting the dependency-scaled matrix ${{D}^{s}}$, and forcing the model to focus on the syntactic dependencies among words.
\begin{gather}
{{{S}}^{l}}=\frac{{{{Q}}^{l}}{{{K}}^{l}}^{^{T}}}{\sqrt{{{d}^{k}}}}\label{eq:3}\\
{{\widetilde{{S}}}^{l}}\text{=}{{{S}}^{l}}\odot {{D}^{s}}\label{eq:4}\\
{{D}_{ij}^{s}}=GaussDist\left( {{d}_{ij}} \right)=\frac{1}{\sqrt{2\pi {{\sigma }^{2}}}}exp\left( -\frac{{{({{d}_{ij}})}^{2}}}{2{{\sigma }^{2}}} \right),\left( i,j \right)=1,...,{I}\label{eq:5},
\end{gather}
where $\widetilde{{S}}_{i}^{l}$ is the $i$-th row of ${{\widetilde{{S}}}^{l}}\in {{\mathbb{R}}^{{IxI}}}$, which represents the scaled alignment score based on the dependency distribution of the $i$-th word ${{x}_{i}}$. $\sigma$ is the Gaussian variance set with respect to a empirical value, usually a tiny number.
Namely, we quantify the dependency closeness among words that are used for re-weighting alignment scores as the Gaussian probability density value. 
Here, ${{d}_{ij}}$ is the dependency distance of the word pair ${{x}_{i}}$ and ${{x}_{j}}$.
Following by this, $GaussDist\left( {{d}_{ij}} \right)$ as the ${{\left( i,j \right)}^{th}}$ entry of the ${{D}_{ij}^{s}}$, which is actually a Gaussian distribution with the variance ${{\sigma }^{\text{2}}}$ and the input ${{d}_{ij}}$.
\begin{gather}
{{Z}^{l}}=softmax ({{\widetilde{S}}^{l}}){{V}^{l}}\label{eq:6}\\
{{O}^{l}}=Concat({{Z}_{1}^{l}},\ldots ,{{Z}_{N}^{l}}){{W}_{O}^{l}}\label{eq:7}.
\end{gather} 

After that, the scaled attention weights are normalized by the softmax function to represent the correspondences between words.
So far, we have achieved that the closer the syntactic dependency to the current encoding word, the higher the attention weight can be assigned.
The attention weight is multiplied by the value vector to produce the output representation of a single attention head. 
Finally, the concatenations of all output representations are passed into a linear projection layer to generate context vectors for subsequent decoding.
In summary, when encoding sentences, the proposed Deps-SAN guides the model to attend to each word in varying scales according to the dependence closeness with the encoding word.

\subsection{Knowledge Sparsing}
\begin{figure*}[!t]
\centering
\includegraphics[width=0.5\textwidth]{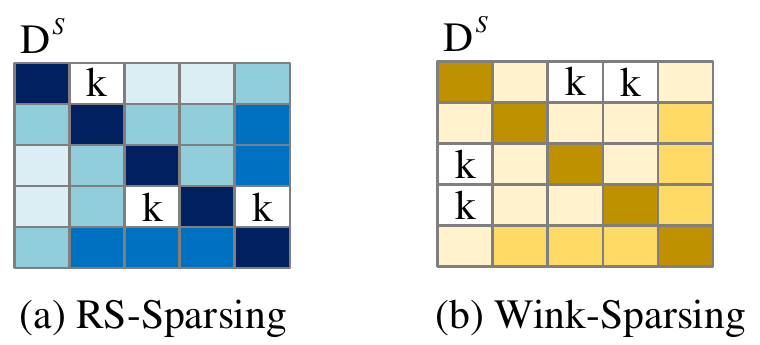}
\caption{The illustration of two knowledge sparsing techniques.} \label{fig:Fig4}
\end{figure*}

In this section, we narrate the motivation for introducing knowledge sparsing and elaborate on two proposed knowledge sparsing techniques in Fig. \ref{fig:Fig4}. 
Since the bilingual corpus does not equip with the parsed result, we are obliged to extract it from an external syntax parser. 
Unfortunately, the parsing accuracy of the syntax parser declines significantly as the distance between the head and dependent increases \cite{mcdonald-nivre-2011-analyzing}. 
To alleviate this obstacle, two knowledge sparsing techniques are provided to confront over-fitting the syntax noise.

\textbf{RS-Sparsing} In this practice, we randomly set each element of ${{D}^{s}}$ to $k$ with probability $q$, as illustrated in Fig. \ref{fig:Fig4}(a). 
We dilute partial syntactic information following the essence of dropout \cite{srivastava2014dropout}.

\textbf{Wink-Sparsing} In this variant, we explicitly concentrate on position whose dependency distance is only within $k$ and masking out other positions. 
This hard clipping mechanism tends to preserving partial syntactic knowledge instead of discarding them, which avert drastic information crash.

\section{Experiment}
\subsection{Setup}
\noindent \textbf{Datasets} 
We evaluated the proposed approach on the widely-used IWSLT14 German-to-English (DE-EN) and WMT14 German-to-English translation tasks. 
For the low-resource IWSLT14 DE-EN translation task, the training set contained 160K sentence pairs\footnote[3]{https://wit3.fbk.eu/archive/2014-01/texts/de/en/de-en.tgz}. 
Following the standard instructions (Cettolo. et al. \cite{cettolo2014report}), we sampled 5$\%$ of the training data for validation and combined multiple test sets IWSLT14.TED.\{dev2010, dev2012,tst2010, tst1011, tst2012\} for testing\footnote[4]{https://github.com/pytorch/fairseq/blob/master/examples/translation/prepare-iwslt14.sh}. 
For the large-scale WMT14 DE-EN translation task, the training set consisted of 4.5M sentence pairs. 
We selected the newstest2015 and newstest2016 as validation sets and test sets respectively. 
We used the raw word rather than the sub-word unit segmented by byte pair encoding algorithm \cite{sennrich2016neural} for better dependency parsing.
That is the reason our Transformer performance is slightly inferior to the reported results of the original paper.

\noindent \textbf{Baseline Systems} 
We train a Transformer model as a robust baseline, and reproduce the following strongly related variants for comparison:
\begin{itemize}
\item[$\bullet$] \textbf{PASCAL\cite{bugliarello-okazaki-2020-enhancing}}: The parent-scaled SAN enable the model to focus on the dependency parent of each token in the encoding phase. 
An auxiliary regularization technique $parent$ $ignoring$ is formulated to avert over-fitting noisy dependencies.

\item[$\bullet$] \textbf{Localness\cite{yang-etal-2018-modeling}}: The localness-aware SAN casts localness modeling as a learnable Gaussian bias to incorporate into SAN.
In this paper, three alternatives of $fixed$, $layer$-$specific$ and $query$-$specific$ localness are proposed according to the local region that need to be paid attention.

\item[$\bullet$] \textbf{Context\cite{yang2019context}}: The context-aware SAN aim to leverage the internal states as context vectors to improve the SAN effect. It is divided into $Global$, $Deep$ and $Deep$-$Global$ three types, according to the diversity of embedded internal state and contextualization.

\item[$\bullet$] \textbf{RPE \cite{shaw-etal-2018-self}}: Shaw et al. extend the SAN by efficiently considering representations of the relative positions, or distance between sequence elements.
\end{itemize}

\noindent \textbf{Settings}
We extract the dependency parsing results from Stanford parser \cite{nivre2016universal}. 
The source and target vocabularies size are both limited to 60K. 
All sentences are limited to 80 words and without aggressive hyphen splitting. 
Case-sensitive 4-gram BLEU \cite{papineni2002bleu} is used as the evaluation metric and paired bootstrap sampling \cite{koehn2004statistical} is applied for statistical significance test\footnote[5]{https://github.com/moses-smt/mosesdecoder/blob/master/scripts/analysis\\/bootstrap-hypothesis-difference-significance.pl}.
We train 80K and 200K steps on the low-resource ISWLT14 dataset and large-scale WMT14 dataset respectively to guarantee that the model has reached convergence.
The hyperparameters $k$ and $q$ for knowledge sparsing are assigned to 6 and 0.1 respectively, which are sufficient to dilute the noisy dependencies of the sentence.
We employ the beam search with a beam size of 5 and length penalty $\alpha=0.6$ for inference. 
Other configurations keep the same as the paper reported by Vaswani et al \cite{vaswani2017attention}. 
All NMT models are trained on an NVIDIA TITAN RTX using \textit{Fairseq} toolkit \cite{ott2019fairseq}.

\subsection{Main Results}
Table. \ref{table_I} shows the translation results of all NMT systems on the IWSLT14 DE-EN and WMT16 DE-EN translation tasks.
Comparing all baselines, we draw the following major conclusions:
\begin{table*}[!t]
\renewcommand{\arraystretch}{0.95} 
\centering
\caption{Translation results of different NMT systems on IWSLT14 DE-EN and WMT16 DE-EN tasks. $\ddagger$/$\dagger$ indicate that the significance of our models is significantly better than that of the Transformer ($p$ $<$ 0.01/0.05). 
``\#Speed'' and ``\#Param'' denote the training speed (seconds/each 100 batches) and the size of model parameters, respectively. 
We highlight the best results in bold for both tasks.}
\label{table_I}
\renewcommand{\arraystretch}{0.9} 
\centering
\begin{tabular}{l||ccc||cc}
\hline
\hline
\multicolumn{1}{c||}{\multirow{2}*{Systems}} & \multicolumn{3}{c||}{IWSLT14 DE-EN} & \multicolumn{2}{c}{WMT16 DE-EN} \\
\cline{2-6} 
& BLEU & \#Speed & \#Param & BLEU & \#Param \\
\hline
\multicolumn{6}{c}{\textit{Reproduced NMT systems of the existing work}}\\  
\hline
Transformer & 29.88 & 23s & 74.85M & 27.48 & 105.57M \\
\quad+PASCAL & 29.75 & 23s & 74.85M & 27.55 & 105.57M \\
\quad+PASCAL+parent ignoring & 29.95 & 24s & 74.85M & 27.73 & 105.57M \\
\quad+Fixed Localness & 28.72 & 24s & 74.87M & 27.21 & 105.59M \\
\quad+Layer-Spec. Localness & 29.89 & 25s & 74.88M & 27.23 & 105.60M \\
\quad+Query-Spec. Localness & 29.63 & 25s & 74.87M & 27.46 & 105.59M \\
\quad+Global Context & 29.92 & 26s & 74.90M & 27.45 & 105.62M \\
\quad+Deep Context & 30.03 & 26s & 74.98M & 27.63 & 105.70M \\
\quad+Deep-Global Context & 30.07 & 26s & 75.03M & 27.55 & 105.75M \\
\quad+RPE & 30.34 & 25s & 74.88M & 27.71 & 105.60M \\
\hline
\multicolumn{6}{c}{\textit{Our NMT systems}}\\  
\hline
\quad+Deps-SAN & 30.44$\ddagger$ & 29s & 74.85M & 27.94$\dagger$ & 105.57M \\
\quad+Deps-SAN+RS-Sparsing & \textbf{30.75}$\ddagger$ & 29s & 74.85M & \textbf{28.21}$\dagger$ & 105.57M \\
\quad+Deps-SAN+Wink-Sparsing & 30.64$\ddagger$ & 29s & 74.85M & 28.13$\dagger$ & 105.57M \\
\hline
\hline 						
\end{tabular}
\end{table*}

\textbf{First}, our model +\emph{Deps-SAN}+\emph{RS-Sparsing} accomplishes the best performance on both translation tasks without additional computation and parameters cost. 
These considerable gains are attributed to the dependency-scaled constraints of Deps-SAN and the anti-syntax-noise ability of RS-Sparsing.

\textbf{Second}, all kinds of localness-aware SANs are inferior to the context-aware SANs. This fact reflects that the context vector directly enhances the sentence semantics compared with implicit localization.
In our implementation, PASCAL does not derive practical improvement from focusing on the dependency parent of the word. 
This probably is related to the word closest to each word is not its dependency parent but its multiple dependency children.
Injecting relative position representation into SAN improves translation performance to a certain extent.

\textbf{Finally}, towards anti-overfitting noisy dependencies, both +\emph{parent} \emph{ignoring} and our knowledge sparsing techniques (i.e. +\emph{RS-Sparsing} and +\emph{Wink-Sparsing} bring a fundamental improvement over both translation tasks. 

\section{Analysis}
\subsection{Ablation Studies}
In this section, we conduct two ablation experiments on the IWSLT14 DE-EN translation task.
First, we investigate the effect of any combination of 
encoder layers equipped with Deps-SAN.
Then, we consider to find the best variance value of Gaussian distribution via a grid search. 
To control the interference of variables, all ablation experiments are implemented on the encoder side solely.

\textbf{Deps-SAN Layer} 
As shown in Table. \ref{table_II(a)}, the performance of the encoder layers plugged with Deps-SAN go up with the increase of layers from bottom to top, until the 3th-layer. 
Our model benefits more from replacing the lower three layers than that of the higher three layers. 
This interesting finding points to the identical conclusion as recent studies \cite{raganato-tiedemann-2018-analysis} --- different layers tend to capture different features.
Further, we infer that the lower layers of the SAN are incline to focus on syntactic information among words, while the higher layers prefer to concentrate on the semantic information of sentence level. 
Based on this impressive insight, we fix the Deps-SAN into the lower three layers to maximize performance.

\textbf{Gaussian Variance} 
From Table. \ref{table_II(b)}, we observe $\sigma=1$ as the optimal Gaussian variance. 
The best results with a variance of 1 benefit from strong supervision of syntactic dependencies by setting a minimum scale. 
Through the bell-shaped curve and non-zero properties, the Gaussian distribution allows the model to focus on the centre of distribution and ensured the integrity of information simultaneously.
\begin{table}[!t]
\setlength{\tabcolsep}{4pt}
\centering
\caption{Validation and test BLEU score for ablation experiments on the IWSLT14 DE-EN translation task.}
\label{table_II}
\renewcommand{\arraystretch}{0.95} 
\centering
\subtable[Combinations of encoder layers]{
\begin{tabular}{c|c|c|c|c}
\hline
\hline
\textbf{Layer} & \textbf{Valid} & \textbf{Test} & \textbf{${{\Delta }_{v}}$} & \textbf{${{\Delta }_{t}}$} \\
\hline 
$[1\text{-}3]$ & 31.42 & 30.44 & - & - \\								
\hline
$[1\text{-}1]$ & 30.71 & 29.83 & -0.71 & -0.63 \\		
$[1\text{-}2]$ & 30.82 & 30.00 & -0.6 & -0.44 \\	
$[1\text{-}4]$ & 30.82 & 29.89 & -0.6 & -0.55 \\		
$[4\text{-}6]$ & 30.41 & 29.46 & -1.01 & -0.98 \\		
$[1\text{-}6]$ & 29.20 & 28.79 & -2.42 & -1.65 \\		
\hline
\hline
\end{tabular}\label{table_II(a)}}
\quad
\subtable[Grid search of Gaussian variance]{\begin{tabular}{c|c|c|c|c}
\hline
\hline
\textbf{Variance} & \textbf{Valid} & \textbf{Test} & \textbf{${{\Delta }_{v}}$} & \textbf{${{\Delta }_{t}}$}\\
\hline
1 & 31.42 & 30.44 & - & - \\
\hline
2 & 30.77 & 29.84 & -0.65 & -0.6 \\		
4 & 30.64 & 29.82 & -0.78 & -0.62 \\	
8 & 30.60 & 30.03 & -0.82 & -0.41 \\		
16 & 30.86 & 30.08 & -0.56 & -0.36 \\		
32 & 30.46 & 29.64 & -0.96 & -0.8 \\		
\hline
\hline
\end{tabular}\label{table_II(b)}}
\end{table}

\subsection{Performance over Sentence Length}
\begin{figure*}[!t]
\centering
{\includegraphics[width=0.6\linewidth]{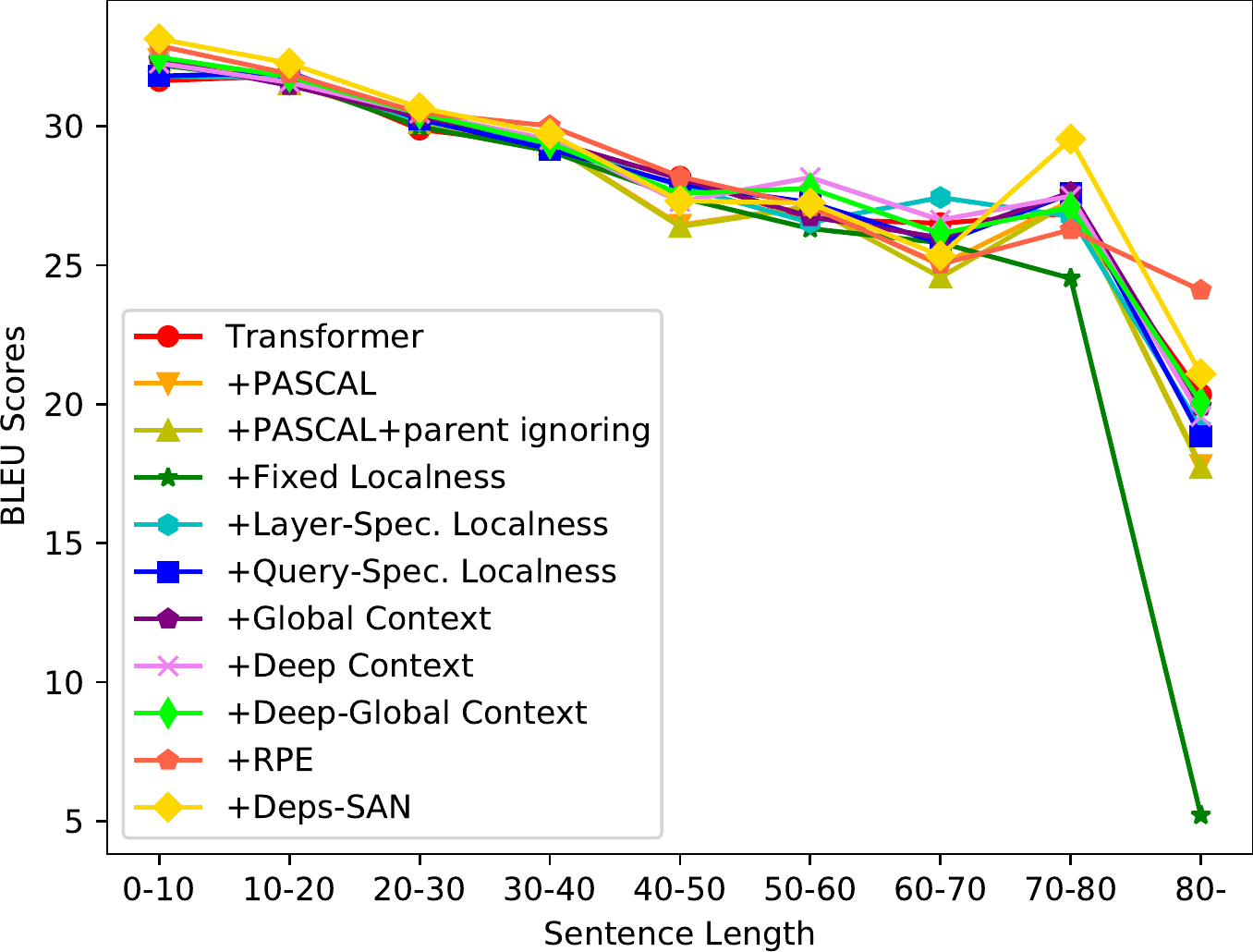} 
\label{fig4}}
\centering
\caption{Performance of different NMT models on test sentences of different lengths.}
\label{fig:Fig5}
\end{figure*}

Following Bahdanau et al. \cite{Bahdanau-Neural}, on the IWSLT14 DE-EN translation task, we divided entire test sentences into 9 disjoint groups according to their lengths\footnote[6]{These groups along with its sentence number are listed below: ([0-10],1657), ([10-20],2637), ([20-30],1381), ([30-40],614), ([40-50],252), ([50-60],122), ([60-70],43), ([70-80],27) and ([80-],17).}.
Fig. \ref{fig:Fig5} presents the performance of varying NMT systems over groups of different sentence lengths. 
We discover that the performance of the proposed method maintains superiority over different sentence lengths.
This success symbolizes our method is available on both long and short sentence translation. 
We are firmly sure this advancement is attributable to explicit syntactic guidance from the Deps-SAN.
The sub-optimal RPE model in the main results is also the second-best in terms of the effect on different sentence lengths.

\subsection{Case Study}
We exhibit the translation results of different NMT models for a test example sentence in Table. \ref{table_III}. 
For this interrogative sentence, the translation mistakes of other models were mainly divided into three categories: sentence pattern errors, misuse of prepositions, and under-translation.
Our model relies on explicit constraints of syntactic dependencies to capture the correct sentence pattern.
The attained accurate translation without any omissions substantiates the reliability of our approach once again.

\begin{table*}[!t]
\centering
\caption{Translation examples generated by different models. Parts of mistake, correct, and under-translation are highlighted in red, blue and underline respectively.}
\label{table_III}
\renewcommand{\arraystretch}{0.95} 
\centering
\begin{tabular}{l||c}
\hline
\hline
\multicolumn{1}{c||}{Source sentence} & wie wäre es, länger bei guter gesundheit zu leben ? \\
\hline
\multicolumn{1}{c||}{Reference sentence} & how about living longer with good health ?\\
\hline
Transformer & \textcolor{red}{what would it be like to} live \textcolor{blue}{on} good health longer ? \\
\quad+PASCAL & \textcolor{red}{what would it be to} live longer with good health ? \\
\quad+PASCAL+parent ignoring & how about living \textcolor{blue}{in} good health longer ? \\
\quad+Fixed Localness & \textcolor{red}{what would it be like to} live longer \textcolor{blue}{on} good health ? \\
\quad+Layer-Spec. Localness & \textcolor{red}{what would it be like to} live longer \textcolor{blue}{on} good health ? \\
\quad+Query-Spec. Localness & \textcolor{red}{what would it be like to} live longer \textcolor{blue}{on} good health ? \\
\quad+Global Context & \textcolor{red}{how would it be to} live longer with good health ? \\
\quad+Deep Context & \underline{how about living longer ?} \\
\quad+Deep-Global Context & how about living longer \textcolor{blue}{for} good health ? \\
\quad+RPE & \textcolor{red}{what would it be like to} live \textcolor{blue}{in} good health longer ? \\
\hline
\quad+Deps-SAN (ours) & \textbf{how about living longer with good health ?} \\
\hline
\end{tabular}
\end{table*}

\section{Conclusion}
This research sheds new light on elevating the Transformer-based syntax-aware NMT.
Specifically, this paper proposes a dependency-scaled self-attention network embedded with the quantified dependency distribution to learn syntactic details and dispel the dispersion of attention distributions.
Two knowledge sparsing techniques are used to avoid our model overfitting the external dependency noises. 
We also investigate the sensitivity of the proposed approach to hyperparameters and their performance over the translation of different sentence lengths and a test example sentence. 
Substantial experimental results and analyses show that our system yields considerable benefits.
In the future, we plan to extend this solution to SAN-based NLP tasks, such as grammar error correction and syntactic parsing tasks.

\section*{Acknowledgment}
This work was supported in part by the National Natural Science Foundation of
China under Grants 62071131, 61771149 and 61772146. 

%
%
\bibliographystyle{splncs04}
\bibliography{llncsref}
\end{document}